\documentclass[10pt,twocolumn,letterpaper]{article}

\usepackage{iccv}
\usepackage{times}
\usepackage{epsfig}
\usepackage{graphicx}
\usepackage{amsmath}
\usepackage{amssymb}

\usepackage{paralist}
\usepackage{booktabs}
\usepackage{xspace} 
\usepackage{stmaryrd}
\usepackage[normalem]{ulem}
\usepackage[tableposition=top]{caption}
\usepackage{appendix}

\usepackage{tabularx}
\makeatletter
\newcommand\notsotiny{\@setfontsize\notsotiny\@vipt\@viipt}
\makeatother

\usepackage{soul} 

\usepackage{color}
\definecolor{crimson}{rgb}{0.86, 0.08, 0.24}
\definecolor{gray}{rgb}{0.5,0.5,0.5}
\definecolor{green}{rgb}{0, 0.4, 0}
\definecolor{orange}{rgb}{1, 0.5, 0}
\definecolor{mahogany}{rgb}{0.75, 0.25, 0.0}
\definecolor{purple}{rgb}{0.6, 0, 0.6}
\definecolor{darkgreen}{rgb}{0, 0.4, 0}
\definecolor{frenchblue}{rgb}{0.0, 0.45, 0.73}
\definecolor{red}{rgb}{1,0,0}
\definecolor{yellow}{rgb}{1,1,0}
\definecolor{magenta}{rgb}{1,0,1}
\definecolor{pink}{rgb}{1,0.412,0.706}
\definecolor{newgreen}{rgb}{0, 0.6, 0.2}

\newcommand{\mycomment}[1]{}
\newcommand{\comment}[1]{}

\newcommand{\tbdegrade}[1]{\color{crimson} #1}

\newcommand{\legendbox}[3]{
    \fboxsep=.7mm \fboxrule=.3mm
    \raisebox{.6mm}{\fcolorbox{black}{#1}{\null}} \hspace{-1.8mm}
}
\definecolor{userOurs}{rgb}{0.27, 0.45, 0.77}
\definecolor{userBLess}{rgb}{0.92, 0.26, 0.21}
\definecolor{userDFill}{rgb}{0.44, 0.68, 0.28}
\definecolor{userNS}{rgb}{0.98, 0.74, 0.02}
\definecolor{userReal}{rgb}{0.65, 0.65, 0.65}
\definecolor{userImg2stylegan}{rgb}{0.86, 0.52, 0.76}

\makeatletter
\DeclareRobustCommand\onedot{\futurelet\@let@token\@onedot}
\def\@onedot{\ifx\@let@token.\else.\null\fi\xspace}

\def\eg{\emph{e.g}\onedot} 
\def\ie{\emph{i.e}\onedot}

\makeatother

\def\eg{e.g.,~}               
\def\ie{i.e.,~}               

\newlength\paramargin
\newlength\figmargin
\newlength\subfigmargin
\newlength\secmargin
\newlength\subsecmargin
\newlength\tabmargin
\newlength\eqmargin

\newcommand{\secref}[1]{Section~\ref{sec:#1}}
\newcommand{\subsecref}[1]{Section~\ref{subsec:#1}}
\newcommand{\figref}[1]{Figure~\ref{fig:#1}} 
\newcommand{\tabref}[1]{Table~\ref{tab:#1}}

\long\def\ignorethis#1{}

\newcommand {\hubert}[1]{{\color{mahogany}\textbf{OuO: }#1}\normalfont}

\usepackage[pagebackref=true,breaklinks=true,letterpaper=true,colorlinks,citecolor=blue,linkcolor=blue,bookmarks=false]{hyperref}

\def\ours{In$\&$Out }
\def\oursc{In$\&$Out-c}

\ifx \submission \undefined
    \newcommand{\sergey}[1]{{\color{purple}{Sergey:#1}}}
	\newcommand{\yenchi}[1]{{\color{frenchblue}{#1}}}

	\newcommand{\todo}[1]{{\color{purple}{(TODO: #1)}}}
	
\else
	\newcommand{\yenchi}[1]{{#1}}

	\newcommand{\todo}[1]{{#1}}
	
\fi

\iccvfinalcopy 

\def\iccvPaperID{2914} 

\ificcvfinal\fi

\begin{document}

\title{\ours: Diverse Image Outpainting via GAN Inversion}
\vspace{-4mm}
\author{Yen-Chi Cheng$^{1}$, Chieh Hubert Lin$^2$, Hsin-Ying Lee$^3$, Jian Ren$^3$, Sergey Tulyakov$^3$, Ming-Hsuan Yang$^{2,4}$ \vspace{-4.5mm}\\ \\
$^1$Carnegie Mellon University\hspace{20pt}$^2$University of California, Merced
\hspace{20pt}$^3$Snap Inc.\hspace{20pt}$^4$Google Research
\\
\small{\url{https://yccyenchicheng.github.io/InOut/}}
}



\twocolumn[{%
\maketitle
\renewcommand\twocolumn[1][]{#1}%
    \vspace{-5mm} 
    \centering
    \includegraphics[width=.93\linewidth]{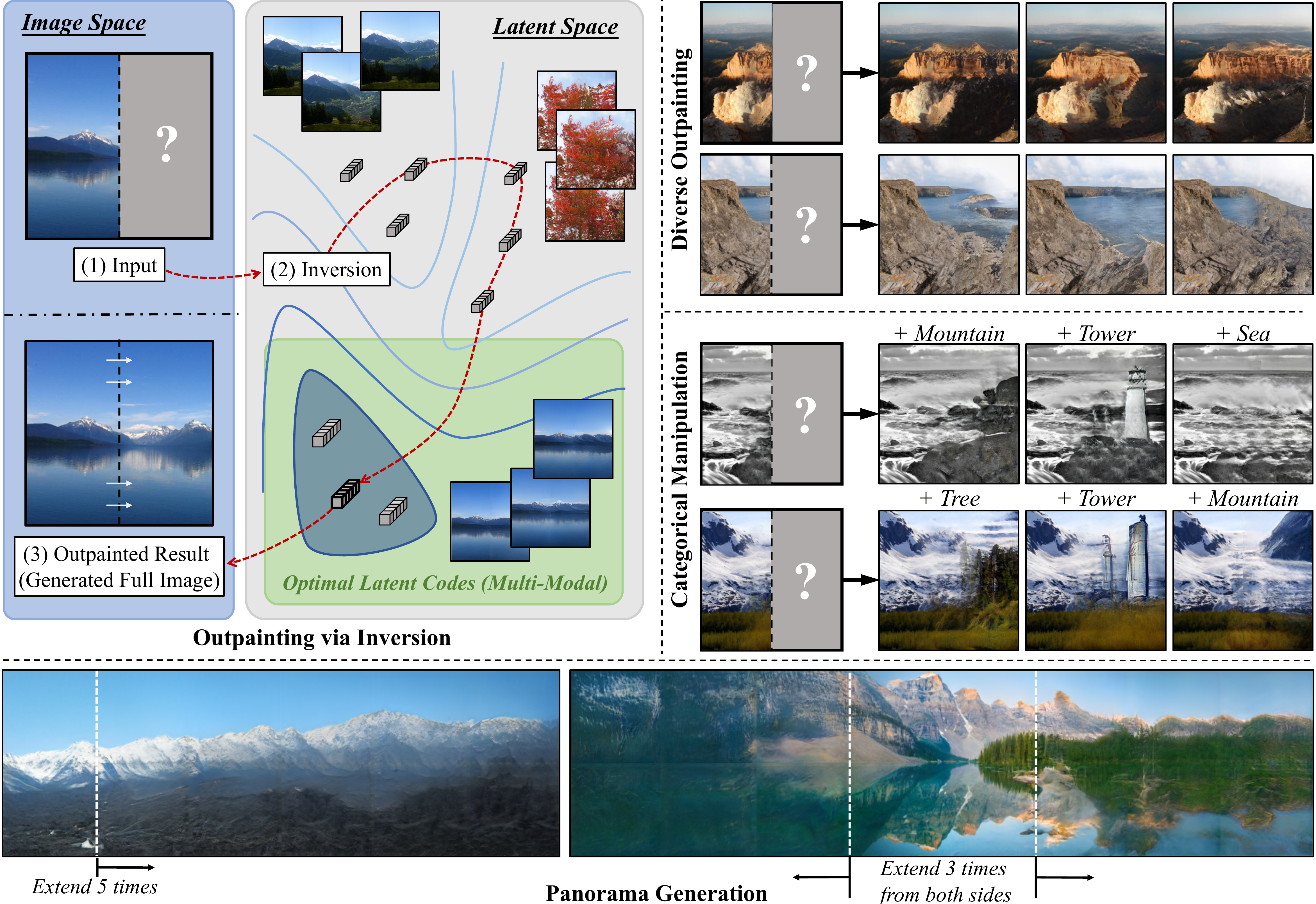}
    \vspace{-2.0mm}
    \captionof{figure}{
        \textit{(top-left)} Given an input image and a trained generator, the proposed algorithm searches for latent codes that can generate images containing the input image.
        We can naturally achieve \textit{(top-right)} diverse image outpainting,
        \textit{(middle-right)} categorical manipulation for outpainting area, and
        \textit{(bottom)} generate panorama with rich and complex structure.
        \label{fig:teaser}
    }
    \vspace{2.5mm} 
}]

\begin{abstract}
\vspace{-4mm}
Image outpainting seeks for a semantically consistent extension of the input image beyond its available content. Compared to inpainting --- filling in missing pixels in a way coherent with the neighboring pixels --- outpainting can be achieved in more diverse ways since the problem is less constrained by the surrounding pixels. 
Existing image outpainting methods pose the problem as a conditional image-to-image translation task, often generating repetitive structures and textures by replicating the content available in the input image. 
In this work, we formulate the problem from the perspective of inverting  generative adversarial networks.
Our generator renders micro-patches conditioned on their joint latent code as well as their individual positions in the image. 
To outpaint an image, we seek for multiple latent codes not only recovering available patches but also synthesizing diverse outpainting by patch-based generation.
This leads to richer structure and content in the outpainted regions. 
Furthermore, our formulation allows for outpainting conditioned on the categorical input, thereby enabling flexible user controls. 
Extensive experimental results demonstrate the proposed method performs favorably against existing in- and outpainting methods, featuring higher visual quality and diversity.
\vspace{-1mm}

\end{abstract}

\vspace{-3mm}
\section{Introduction}
\vspace{\secmargin}
Given an input image, we can easily picture how adjacent images might look, had they been captured. 
For example, given an image of mountains, we can picture the surroundings covered by forests or snow, imagine a lake beneath the hillside, and visualize cliffs near the ocean. 
This mental skill depends on our prior experience and exposure to diverse scenery. 
In other words, this is an \textit{image outpainting} task. 
It can enable various content creation applications such as image editing using extrapolated regions, panorama image generation, and extended immersive experience in virtual reality, to name a few. 

Recent advances in image inpainting~\cite{liu2018image,pathak2016context,yu2018generative,yu2018free} do not directly address the outpainting problem as the former has more context to deal with --- the missing pixels have a larger amount of available surrounding pixels, serving as the boundary conditions and providing crucial guidance for inpainting. %
In contrast, the outpainting problem can rely only on the context of the available image, with only a scarce number of pixels near the boundary available as the boundary condition. 
Furthermore, the texture and semantics of the outpainted regions should be coherent with that of the input. 
Finally, outpainting methods ought to support diversity in the generated content. 
A similar analogy is between video interpolation and video prediction, where the former deals with existing events~\cite{jiang2018super} while the latter tries to model multiple futures~\cite{wichers2018hierarchical}.

In the literature, image outpainting is addressed from the image-to-image translation (I2I) perspective~\cite{teterwak2019boundless,yang2019very}.
These methods aim to learn a deterministic mapping from the domain of partial images to the domain of complete outpainted images. 
%
This formulation is limited in several respects. 
First, for the I2I methods, the available pixels serve as a strong source of context, thereby facilitating leakage of textures and structures of the input to the output and  leading to the repetitive nature of the outpainting (as shown in panorama results in~\cite{teterwak2019boundless}). 
Second, existing I2I-based methods are deterministic~\cite{teterwak2019boundless,yang2019very}, while in reality there exist numerous ways each image can be outpainted. 
Applying the available multimodal I2I methods~\cite{huang2018munit,DRIT} to the outpainting problem is non-trivial.

In this work, we propose \ours, a framework that tackles the outpainting problem by inverting generative adversarial networks (GANs)
~\cite{abdal2019image2stylegan,bau2020semantic,creswell2018inverting,ma2018invertibility,zhu2020domain}. 
%
Similar to Lin et al.~\cite{lin2019coco}, we extend a StyleGAN2-based~\cite{Karras2019stylegan2} generator to perform generation in a coordinate conditional manner and independently generate spatially consistent micro-patches. %
Each micro-patch shares the global latent code with the rest of micro-patches in the image, while having a unique coordinate label. 
Outpainting can then be formulated as finding the optimal latent codes for the available input micro-patches, followed by generating the desired regions by providing the proper coordinate conditioning. 
To search for the latent code, we propose a GAN inversion process that finds multiple latent codes producing diverse outpainted regions, unlocking diversity in the output. 
In addition, we propose a categorical generation schema to enable flexible user control. 
\figref{teaser} shows examples of multi-modal and categorical outpainting.

We qualitatively and quantitatively evaluate the proposed method on the Place365 \cite{zhou2017places} dataset, and the Flickr-Scenery dataset which we collected. 
%
We leverage Fr\'echet Inception Distance (FID)~\cite{fid} and conduct a user study to evaluate the realism of outpainted images.
Since the proposed method can achieve multi-modal generation, we measure the diversity using the Learned Perceptual Image Patch Similarity (LPIPS) metric~\cite{zhang2018lpips}.
Finally, we demonstrate the scenario of categorical generation in the outpainting area and the panorama generation.

\vskip \secmargin
\section{Related Work}
\vspace{\secmargin}

\vspace{2mm}
\noindent\textbf{Generative Adversarial Networks.}
Generative models aim to model and sample from a target distribution.
Generative adversarial networks~\cite{goodfellow2014generative}, among various generative models, have demonstrated superior performance in generating high-quality samples.
The core idea of GANs is a two-player game between a generator aiming to map noise vectors to realistic images and a discriminator attempting to discriminate the generated images from the real ones.
GANs facilitate a variety of creation tasks such as image-to-image translation~\cite{CycleGAN2017}, text-to-image generation~\cite{tseng2020retrievegan,han2017stackgan}, semantic image synthesis~\cite{cheng2020segvae,park2019SPADE}, video generation~\cite{li2018video,p2pvg2019}, etc.
However, most of the models generate new images from scratch given various conditional contexts, and generally lack the ability to perform editing and interactive manipulation on existing images.

\begin{figure*}[t!]
    \centering
    \includegraphics[width=\linewidth]{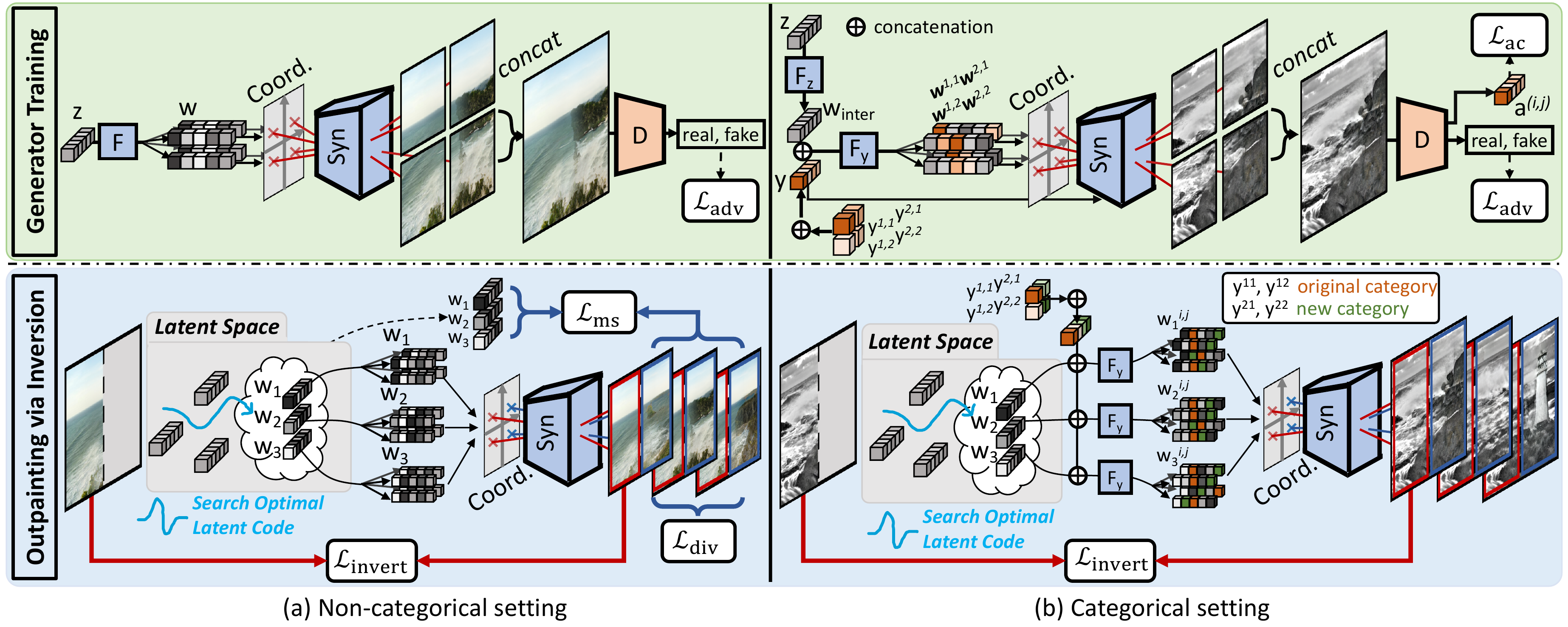}
    \vspace{-6mm}
    \caption{
    \textbf{Overview.} 
    \textbf{Generator training:} The proposed generator adopts StyleGAN2 as the backbone and incorporates coordinate conditioning, where micro-patches are generated conditioned on their positions in the image.
    \textbf{Outpainting via inversion:} We search for latent codes that can recover the given partial images and synthesize diverse samples in the outpainted regions.
    In addition to the \textit{unconditional setting}, we introduce a \textit{categorical setting} variant, which enables flexible user controls for the categorical manipulation on each outpainting micro-patch.
    }
    \vspace \figmargin
    \label{fig:overview}
\end{figure*} 

\vspace{2mm}
\noindent\textbf{GAN Inversion.}
To fully exploit the ability and explore the interpretability of well-trained GANs, GAN inversion has been proposed to find the latent codes that can accurately recover given images for a trained GAN model.
There are two main branches of approaches.
\textit{Encoder-based methods}~\cite{brock2016neural,dumoulin2016adversarially,perarnau2016invertible} adopt an additional encoder to learn the mapping from the image domain to the latent space.
\textit{Optimization-based methods}~\cite{abdal2019image2stylegan,abdal2020image2stylegan++,creswell2018inverting,lipton2017precise,ma2018invertibility} use gradient-based optimization methods (\ie stochastic gradient decent and ADAM) with reconstruction loss as the objective function to find latent codes that can recover input images.
Other variants use encoders to get an initialization for the optimization process~\cite{bau2020semantic,zhu2016generative}, or modify the training framework by incorporating invertibility~\cite{donahue2016adversarial,zhu2020domain}.
In this work, we adopt the optimization-based technique to tackle the image outpainting task. 

\vspace{2mm}
\noindent\textbf{Image Inpainting.}
From the aspect of filling missing pixels in images with generative models, the inpainting problem is conceptually related to the outpainting task.
Existing image inpainting methods can be categorized into two groups.
The first line of work leverages patch similarity and diffusion to obtain essential information from known regions~\cite{efros1999texture,yang2019very}.
These methods usually work well on textures but fail to learn semantic structures of images.
The other line of work adopts a learning-based approach to gain better semantic understanding~\cite{liu2018image,pathak2016context,yu2018generative,yu2018free}.
Most methods apply an encoder-decoder model with the reconstruction loss and adversarial loss to ensure the filled content is smooth and realistic.
In this work, we focus on image outpainting instead of inpainting.
Image outpainting is more challenging since it entails \textit{creating} new contents rather than \textit{filling in} partial regions, requiring a substantial understanding of scenes. 

\vspace{2mm}
\noindent\textbf{Image Outpainting.}
Most image outpainting methods~\cite{efros1999texture,kopf2012quality,sivic2008creating,wang2014biggerpicture} apply patch-based retrieval and matching algorithms to predict possible extrapolation.
Recently, several approaches~\cite{teterwak2019boundless,wang2019wide,yang2019very} apply GAN models and formulate the problem as an image-to-image translation task.
However, the conditional formulation relies heavily on the given available pixels and tends to create repetitive textures and structures.
%
To the best of our knowledge, the proposed method is the first attempt to tackle the image outpainting task from the GAN inversion perspective.
\vspace{-2mm}
\section{Diverse Outpainting via Inversion}
\vspace{\secmargin}
\label{sec:method}

\vspace{0mm}
\noindent\textbf{Overview.}
The goal of image outpainting is to outward-synthesize unknown regions with respect to the given input image.
The proposed \ours consists of two stages, \textbf{\emph{generator training}} and \textbf{\emph{outpainting via inversion}}.
In \subsecref{coco}, we first introduce a generator based on the StyleGAN~\cite{karras2019style,Karras2019stylegan2} and COCO-GAN~\cite{lin2019coco}.
It is trained to output micro-patches conditioned on the joint latent and on the coordinate of the patch in the output image. We do not specifically optimize the generator to perform outpainting.
During the outpainting stage (\subsecref{inout}) we find the optimal latent code for the available input patches in the latent space of the trained patch-based generator. Outpainting is then performed by combining the desired coordinates and the found optimal latent code and generating a new patch. We further propose a categorical-conditioning scheme to enable controllable outpainting.
Finally, a simple blending algorithm to further mitigate the artifacts is introduced in \subsecref{blending}.
\vspace{-4mm}
\subsection{Coordinate-conditioned Generator}
\vspace{\subsecmargin}
\label{subsec:coco}
In this work, we handle two different settings:
%
(a)  \textit{non-categorical generation} that synthesizes images from latent codes, and
(b) \textit{categorical generation} that uses categorical labels as additional conditional context, enabling more user-control in the following inversion stage.

\vspace{2mm}
\noindent \textbf{Non-categorical generation.}
We use the StyleGAN2~\cite{Karras2019stylegan2} as our backbone architecture.
Given a latent $\mathbf{z}$ from the input latent space $\mathcal{Z}$, we obtain an intermediate code $\mathbf{w} \in \mathcal{W}$ by a non-linear mapping network $F$. 
Similar to in~\cite{wulff2020improving},
we map $\mathbf{w}$ to a Gaussianized space $\mathcal{V}$.
The mapping is achieved via a Leaky ReLU (LRU) with a negative slope of $5$, that is, $\mathbf{v} = \mathrm{LRU}_{5.0}(\mathbf{w})$.
The outpainting quality in the later GAN inversion stage can be substantially improved with the additional Gaussianized space.
The necessity of adopting the Gaussianized space is discussed in \subsecref{inout}.

We formulate the image outpainting problem as finding the latent codes that synthesize images overlapping with the input image. 
In the inversion process, we seek for a latent code for the whole image while having only a part of the image available. 
%
%
Therefore, instead of generating a full image, the generator synthesizes several \textit{micro-patches} $\{I^{i,j}_{\textrm{micro}}\}_{i,j=1,\dots,n}$, which will be concatenated to form a full image $I_f$.
Each patch depends on the joint latent code and its coordinates.
%
For an $n \times n$ micro-patches generation setting, the corresponding coordinates to $\{I_{\mathrm{micro}}^{i,j}\}_{i,j=1,\dots,n}$ are $\{c^{i,j}\}_{i,j=1,\dots,n}$.
We set $c^{1,1} = (-1,-1)$, $c^{n,n} = (1,1)$, and the rest, if any, are obtained by linear interpolation.
The output image $I_f$ is generated as follows:
\vspace{\eqmargin}
\begin{equation}
\begin{aligned}
    \mathbf{w} &= F(\mathbf{z}) \, , \\
    \mathbf{v} &= \mathrm{LRU}_{5.0}(\mathbf{w}) \, , \\
    I_{\mathrm{micro}}^{i,j} &= G(\mathbf{v}, c^{i, j}) \, , \\
    I_{\mathrm{f}} &= \underset{i,j=1,\dots,n}{\mathrm{concat}}({I_{\mathrm{micro}}^{i,j}}) \, .
\end{aligned}
\vspace{\eqmargin}
\end{equation}
We train the generator using the Wasserstein-GAN loss~\cite{arjovsky2017wasserstein} with real full-images $I_{\mathrm{r}}$ and generated full-images $I_{\mathrm{f}}$:
\vspace{\eqmargin}
\vspace{1mm}
\begin{equation}
    \mathcal{L}_{\mathrm{adv}} = \textstyle  \mathbb{E}_{I_{\mathrm{r}}} \, [ \, D(I_{\mathrm{r}}) \, ] - \textstyle  \mathbb{E}_{z} \, [ \, D(I_{\mathrm{f}}) \, ] \, .
\vspace{\eqmargin}
\vspace{1mm}
\end{equation}

\noindent \textbf{Categorical generation.}
To enable fine-grained user control in the inversion stage, we propose a categorical generation schema.
Given a real image $I_{\mathrm{r}}$, we divide it into micro-patches $\{I_{\mathrm{micro}}^{i,j}\}$.
We then obtain the categorical labels $\{y^{i,j}\}$ for each micro-patch using the off-the-shelf DeepLabV3~\cite{chen2017rethinking} model, and set the $k$-th element in the multi-class binary label vector $y^{i,j}$ to $1$ if any of the pixels within $I_{\mathrm{micro}}^{i,j}$ is recognized as the $k$-th class.

To use categorical information as a conditional input, we split the nonlinear mapping network $F$ into $\{F_z, F_y\}$.
Here, $F_z$ operates the same way as $F$ in the non-categorical setting, while $F_y$ takes $\{y^{i,j}\}$ as an additional input and fuses the information with the output of $F_z$.
The new $w_{i,j}$ under the categorical setting is computed by
\vspace{\eqmargin}
\begin{equation}
\begin{aligned}
    \mathbf{w}_{\mathrm{inter}} &= F_z(\mathbf{z}) \, , \\
    \mathbf{w}^{i, j} &= F_y(\mathbf{w}_{\mathrm{inter}}, y^{i,j}) \, .
\end{aligned}
\vspace{\eqmargin}
\end{equation}
Next, similar to the non-categorical generation setting, we first Gaussianize the code with $\mathbf{v}^{i,j} = \mathrm{LRU}_{5.0}(\mathbf{w}^{i,j})$, generate micro-patches $I_{\mathrm{micro}}^{i,j} = G(\mathbf{v}, c_{i, j})$, then concatenate $\{I_{\mathrm{micro}}^{i,j}\}_{i,j=1,\dots,n}$ into a full image $I_{\mathrm{f}}$.
We apply 
an auxiliary classifier~\cite{odena2017conditional} that uses the last intermediate features of $D$ to perform multi-class classification $a^{i,j}$ for all ${I_{\mathrm{micro}}^{i,j}}$, which aims at learning a proper conditional distribution of ${I_{\mathrm{micro}}^{i,j}}$ regarding the $y^{i,j}$ input to $G$.
\vspace{\eqmargin}
\begin{equation}
\begin{aligned}
   \mathcal{L}_{\mathrm{adv}} &= \textstyle \mathbb{E}_{I_{\mathrm{r}}} \, [ \, D(I_{\mathrm{r}}) \, ] - \textstyle \mathbb{E}_{z} \,  [ \, D(I_{\mathrm{f}}) \, ] \, , \\
   \mathcal{L}_{\mathrm{{cls}}} &= \mathrm{BCE}(a^{i,j}, y^{i,j}) \, , 
\end{aligned}
\vspace{\eqmargin}
\end{equation}
where $\mathrm{BCE}$ is the binary cross entropy loss function. The full training objective is:
$$
\underset{D}{\mathrm{min}} \,\, \underset{G}{\mathrm{max}} \,\, \mathcal{L}_{\mathrm{adv}} \, + \, \underset{G,D}{\mathrm{min}} \,\, \mathcal{L}_{\mathrm{cls}} \, .
$$

\vspace{-1mm}
\subsection{GAN Inversion with Diversity Loss}
\vspace{\subsecmargin}
\label{subsec:inout}
Given a trained coordinate-conditioned generator $G$ as discussed in the previous section and an input image $R$ as a reference, we generate a set of possible outpainted images by composing $R$ with generated micro-patches $\{O^m\}$.
For brevity and notation clarity, here we assume $G$ is trained with a grid of $2 \times 2$ micro-patches, $\{ R_\mathrm{micro}^{1, 1}, R_\mathrm{micro}^{1, 2}, R_\mathrm{micro}^{2, 1}, R_\mathrm{micro}^{2, 2} \}$. 
Furthermore, for presentation simplicity, we assume $R$ to be on the left and consists of two left-side micro-patches (\ie $R_\mathrm{micro}^{1, 1}$ and $R_\mathrm{micro}^{1, 2}$), while the outpainted area $\{O^m\}$ on the right, as shown in the lower-half of \figref{overview}.
Note that, in practice, $G$ is not restricted to $2 \times 2$, $R$ can be of any resolution, and outpainting can be performed using an arbitrary direction.

Similar to existing optimization-based GAN inversion methods, we seek for the optimal latent code $w$ that recovers the input image.
The basic loss function is:
\vspace{\eqmargin}
\begin{equation}
\begin{aligned}
    R_f &= \mathrm{concat}(G(\mathbf{v}, c^{1,1}), G(\mathbf{v}, c^{1,2})) \, , \\
    \mathcal{L}_{\mathrm{mse}} &= \lVert R - R_f  \lVert_{2} \, , \\ 
    \mathcal{L}_{\mathrm{percept}} &= \mathrm{Percept}(R, R_f) \, , 
\end{aligned}
\vspace{\eqmargin}
\end{equation}
where  $\mathbf{v} = \mathrm{LRU}_{5.0}(\mathbf{w})$ and $\mathrm{Percept}$ is the perceptual distance proposed in~\cite{zhang2018lpips}.

The outpainting process requires not only the reconstructed parts to be correct $R$ (\ie  $I_{\mathrm{micro}}^{1, 1}$ and $I_{\mathrm{micro}}^{1, 2}$), but the outpainted parts (\ie  $I_{\mathrm{micro}}^{2, 1}$ and $I_{\mathrm{micro}}^{2, 2}$) to be realistic and consistent.
Note that the continuity and consistency between micro-patches are enforced by joint latent and the coordinate conditioning schema. During the generator training, the latent is sampled from a Gaussian distribution.
%
Therefore, it is crucial to encourage the sought latent code $\mathbf{w}$ to belong to the domain of the training data, and be interpretable by $G$, instead of overfitting to the given image with the out-of-domain latent code. 
As the first step, in \subsecref{coco}, we add an additional \textit{Gaussianized} space~\cite{wulff2020improving} $\mathcal{V}$ after $\mathcal{W}$, simplifying the complex and arbitrarily shaped $\mathcal{W}$ with $\mathrm{LRU}_{5.0}$.
Next, with the Gaussianized $\mathcal{V}$, we can easily derive the mean $\mathbf{\mu}$ and covariance matrix $\mathbf{\Sigma}$ of the distribution $p(\mathbf{v})$, where $\mathbf{v} \in \mathcal{V}$.
We encourage recovered $\mathbf{v}$ to be in the training distribution by regularizing its prior:
\begin{equation}
\mathcal{L}_{\mathrm{prior}} = {(\mathbf{v}-\mathbf{\mu})}^\top\mathbf{\Sigma}^{-1}(\mathbf{v}-\mathbf{\mu}) \, .
\end{equation}

To enable diverse outpainting, we apply two different objective functions.
Assuming we target at generating $m$ different outpainted results, we first explicitly penalize the inverted latent codes with their pairwise distance:
\vspace{\eqmargin}
\begin{equation} 
  \mathcal{L}_\mathrm{div} = - \sum_{i=1}^m\sum_{j=i+1}^{m} \lVert \mathbf{w}^i - \mathbf{w}^j \lVert_{1} \, .
\vspace{\eqmargin}
\end{equation}
Then, to further encourage the model to seek for different final latent codes within the latent space, we apply a mode-seeking regularization~\cite{MSGAN}:
\vspace{\eqmargin}
\begin{equation}
  \mathcal{L}_\mathrm{ms} = \sum_{i=1}^m\sum_{j=i+1}^{m} (\frac{
        \lVert G(\mathbf{w}^i) - G(\mathbf{w}^j) \rVert_1
    }{
        \lVert \mathbf{w}^i - \mathbf{w}^j \rVert_1
    }) \, .
\vspace{\eqmargin}
\end{equation}
%
The full objective of our optimization-based inversion is:
\vspace{\eqmargin}
\begin{equation}
\begin{aligned} \label{eq:total-loss}
    \mathop{\arg\min}_{\{\mathbf{w_i}\}\in \mathcal{W}} \quad
    & \lambda_{\mathrm{mse}}\mathcal{L}_{\mathrm{mse}} + \lambda_{\mathrm{percept}}\mathcal{L}_{\mathrm{percept}} + \\
    & \quad \lambda_{\mathrm{prior}}\mathcal{L}_{\mathrm{prior}} + \lambda_{\mathrm{div}}\mathcal{L}_{\mathrm{div}}+ \lambda_{\mathrm{ms}}\mathcal{L}_{\mathrm{ms}} \, ,
\end{aligned}
\vspace{\eqmargin}
\end{equation}
where the hyper-parameters $\lambda$'s control the importance of each term.
Note that such an inversion paradigm is the same for both non-categorical and categorical settings, except the categorical setting seeks for $\mathbf{w}_{\mathrm{inter}}$ instead of $\mathbf{w}$.

\vspace{-1mm}
\subsection{Patch Blending}
\vspace{\subsecmargin}
\label{subsec:blending}

%
As described in \subsecref{inout}, the inversion process requires the reconstruction of the given parts and the prediction of the outpainting parts, where the continuity and consistency are enforced in the training stage.
However, even with the help of the prior loss $\mathcal{L}_{\mathrm{prior}}$, the outpainting occasionally leads to tiny seams between patches after concatenating patches.
%
Due to simple merging of patches, the outpainted images are likely to contain artifacts. 
As such, we introduce an image blending method to address this issue. 
In addition to the reference image $R$ and outpainted area $O$, we generate the patches located halfway between $R$ and $O$. 
Take  $R = I_{\mathrm{micro}}^{1, 1},  I_{\mathrm{micro}}^{1, 2}$ and  $O = I_{\mathrm{micro}}^{2, 1}, I_{\mathrm{micro}}^{2, 2}$ as an example, the additional region $A$ is generated with coordinate $(0, -1)$ and $(0, 1)$. 
We then linearly blend the overlapped area between $R$ and  $A$ and the area between $O$ and $A$. 
%

Despite its simplicity, this post-processing step provides sufficient quality for our purpose. 
In practice, we observe that the generator can accurately interpolate the positions of the extended silhouette of landscapes with respect to the coordinate interpolation. 
The only rare artifacts of blending that occur in practice are in the categorical setting when a large foreground object (\eg tower) is rendered from patches with slight ghosting artifacts.



\vspace{-0mm}
\section{Experimental Results}
\vspace{1mm}
\vspace{\secmargin}
\begin{table}[t]
\caption{
    \textbf{Quantitative comparisons.} 
   The proposed method outperforms related state-of-the-art baselines on both Places365 and Flickr-Scenery datasets in FID and IS metrics, measuring both the visual quality and diversity.
    }
    \vspace{-3mm}
\centering
\small
\begin{tabular}{l  cc cc} 
    \toprule
     & \multicolumn{2}{c}{Place365} & \multicolumn{2}{c}{Flickr-Scenery} \\
    \cmidrule(r){2-3} \cmidrule(r){4-5}
    {Method} & {FID $\shortdownarrow$ } & {IS $\shortuparrow$ }  & {FID $\shortdownarrow$ } &{IS $\shortuparrow$ }  \\
    \midrule
    Boundless~\cite{teterwak2019boundless}  & $35.02$ & $6.15$ & $61.98$ & $6.98$ \\
    NS-outpaint~\cite{yang2019very} & $50.68$ & $4.70$ & $61.16$ & $4.76$ \\
    DeepFillv2~\cite{yu2018generative,yu2018free}   & $56.14$ & $5.69$ & $62.47$ & $5.38$ \\
    Image2StyleGAN++~\cite{abdal2020image2stylegan++} & \underline{25.36}&6.71&40.39& 7.10 \\
    \midrule
    \ours (Ours)  & \textbf{23.57} & \underline{7.18} & \textbf{30.34} & \textbf{7.16} \\
    \oursc (Ours) & 29.24 & \textbf{7.69} & \underline{33.17} & \underline{7.15}\\
    \bottomrule
\end{tabular}
\vspace{\tabmargin}
\label{tab:quan_fid_div}
\end{table}

\begin{figure}[t]
    \centering
    \setlength{\tabcolsep}{0em}
    \begin{tabular}{c c}
        \includegraphics[width=0.47\linewidth]{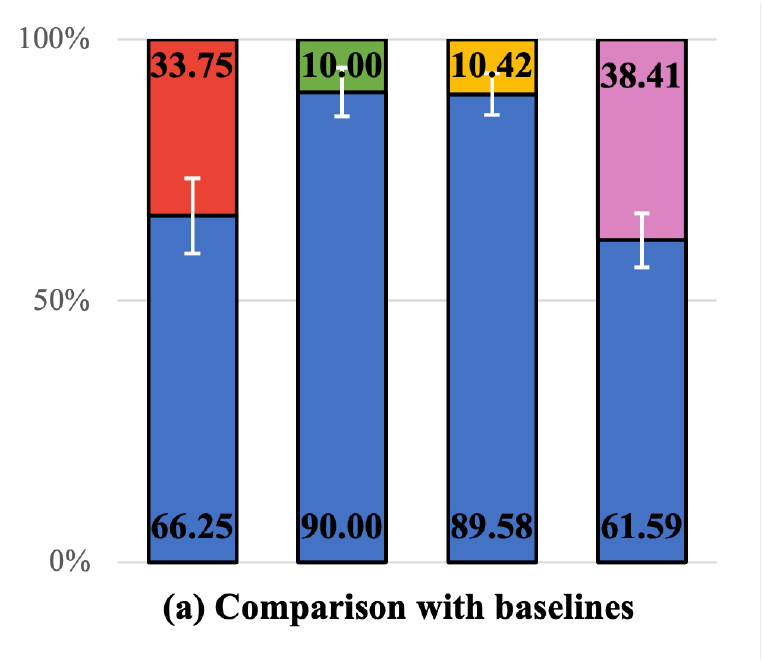}
        \hfill & \hfill
        \includegraphics[width=0.47\linewidth]{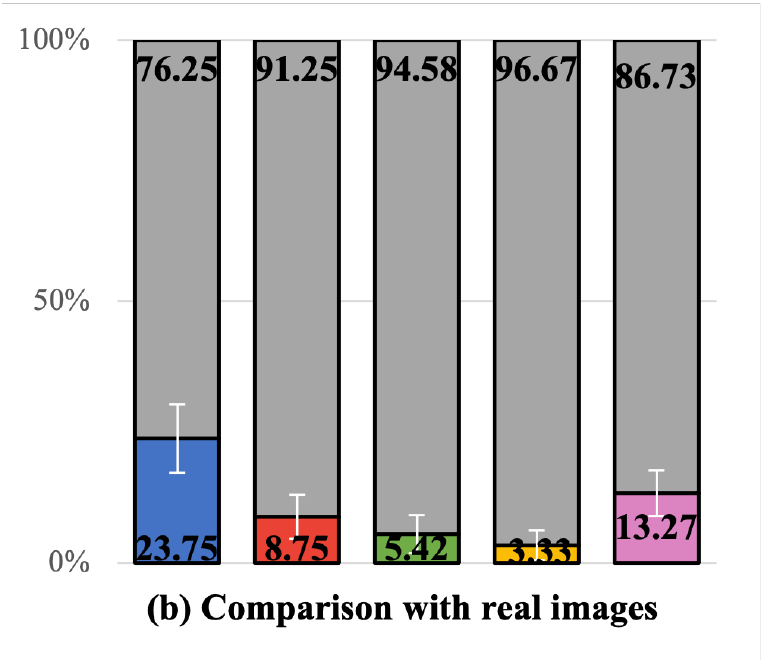} \\
    \end{tabular}  
    \setlength{\tabcolsep}{0.5em}
    \hfill
    \fbox{\begin{minipage}[t]{0.95\linewidth}
    \begin{tabularx}{\linewidth}{l c c c c}
        \raggedleft
        \legendbox{userOurs} \hspace{0mm} \notsotiny In\&Out &
        \legendbox{userBLess} \hspace{0mm} \notsotiny Boundless & 
        \legendbox{userDFill} \hspace{0mm} \notsotiny DeepFillv2 &
        \legendbox{userNS} \hspace{0mm} \notsotiny NS-outpaint \\
        \hfill
        \legendbox{userImg2stylegan} \hspace{0mm} \notsotiny Image2StyleGAN++ &
        \legendbox{userReal} \hspace{0mm} \notsotiny Real Data 
    \end{tabularx}
    \end{minipage}}
    \hfill
    \vspace{-3mm}
    \caption{
    \textbf{User Studies.}
    We conduct user studies to quantify the visual quality in two settings: (a) comparison with baselines, and (b) comparison with real images.
    We mark the $95\%$ confidence interval with white bars. 
    }
    \vspace \figmargin
    \label{fig:userstudy}
\end{figure}

\begin{table}[t]
\caption{
    \textbf{Ablation studies}.
    We show the necessity of each component using FID and LPIPS for quality and diversity.
    %
}
\vspace{-2mm}
\centering
\setlength{\tabcolsep}{0.25em}
\small
\begin{tabular}{l cc cc} 
    \toprule
    {$\#$ output}& \multicolumn{2}{c}{m=2} &\multicolumn{2}{c}{m=3} \\ 
        \cmidrule(r){2-3} \cmidrule(r){4-5}
    Method & \small{FID $\shortdownarrow$} & \small{Diversity $\shortuparrow$} & \small{FID $\shortdownarrow$} & \small{Diversity $\shortuparrow$}\\

    \midrule
    \ours w/o $\mathcal{L}_{\mathrm{div}}$, $\mathcal{L}_{\mathrm{ms}}$ & 30.12 & \tbdegrade{0.183} & 30.28  & \tbdegrade{0.176} \\
    \ours w/o $\mathcal{L}_{\mathrm{ms}}$   & 29.85  & 0.201  &29.80 & 0.206 \\
    \ours w/o $\mathcal{L}_{\mathrm{div}}$   & 29.75  & 0.204  &  33.97 &  0.201 \\
    \cmidrule(lr){1-5}
    \ours w/o $\mathcal{L}_{\mathrm{prior}}$  & \tbdegrade{36.56}  & 0.216 &  \tbdegrade{36.53}  & 0.220\\
    \cmidrule(lr){1-5}
    \ours  & 30.26 & 0.211 &  30.18 & 0.223 \\
    \bottomrule
\end{tabular}
\vspace{\tabmargin}
\vspace{-2mm}
\label{tab:ablation}
\end{table}

\begin{figure*}[t!]
    \centering
    \includegraphics[width=.97\linewidth]{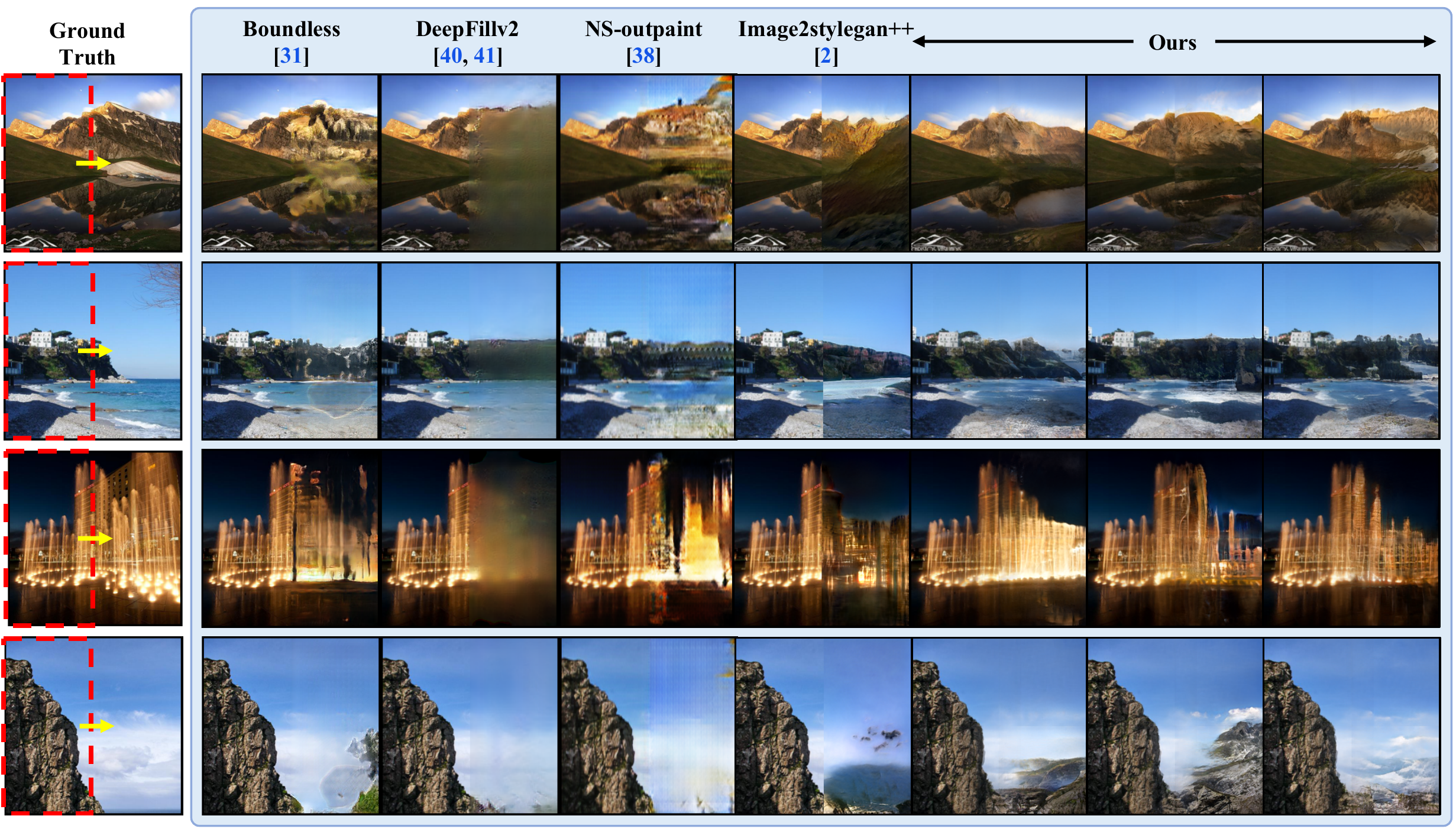}
    \vspace{-2mm}
    \caption{
    \textbf{Comparison to related work.}
    The qualitative comparison against other related methods show that the proposed approach is more stable, synthesizes richer context with more complex structures, and is able to handle some of the difficult complex scenes.
    (The input regions to all methods are marked with red dashes.)
    }
    \vspace \figmargin
    \label{fig:comparison}
\end{figure*}

\vspace{1mm}
\noindent \textbf{Dataset.} We evaluate our method on scenery datasets
since they are the most representative and natural use-cases of outpainting.  
We perform experiments on the Places365~\cite{zhou2017places} dataset and a collected Flickr-Scenery dataset. 
More results on images with structured samples (\eg buildings) could be found in the supplementary material.
%
Similar to \cite{teterwak2019boundless}, we evaluate 
our method on a subset of the Places365 dataset. 
%
We select $25$ scenery classes from the Places365 dataset with a subset of $62{,}500$ samples. 
To further analyze the generalization of our method, we construct a Flickr-Scenery dataset by collecting a large-scale scenery image database of $54{,}710$ images from Flickr.  
All images are center-cropped and resized to 256$\times$256 pixels. 
For both datasets, we split the data into $80\%$, $10\%$, $10\%$ for training, validation, and testing. 
All quantitative and qualitative experiments are evaluated on the testing set only.
The  source code, trained models, and Flickr-Scenery dataset will be made publicly available. 


\vspace{-2mm}
\paragraph{Hyperparameters.} We use the default setup and parameters from the StyleGAN2~\cite{Karras2019stylegan2}, including architecture, losses, optimizer, use of lazy regularizer, and implementation of inversion pipeline. 
The hyperparameters of our model are set as follows:
\begin{compactitem}
    \item \textbf{Generator training.} Following the example in \subsecref{inout}, we train our generator with a grid of $2 \times 2$ micro-patches, and different from COCO-GAN~\cite{lin2019coco}, we train the discriminator 
    with full images.  
    \item \textbf{Outpainting via inversion.} The weighting factors in Equation~\ref{eq:total-loss} are: $\lambda_{\mathrm{mse}}=0.01$, $\lambda_{\mathrm{percept}}=1$, $\lambda_{\mathrm{prior}}=0.001$, $\lambda_{\mathrm{div}}=0.001$, and $\lambda_{\mathrm{ms}}=0.001$.
\end{compactitem}


\vspace{-2mm}
\paragraph{Evaluated Methods.}
We carry out quantitative and qualitative experiments with the state-of-the-art image outpainting methods (Boundless~\cite{teterwak2019boundless} and NS-outpaint~\cite{yang2019very}) and also  image-inpainting methods (DeepFillv2~\cite{yu2018generative,yu2018free} and  Image2stylegan++~\cite{abdal2020image2stylegan++}). \oursc~is the proposed method with categorical setting.


\begin{figure*}[t!]
    \centering
    \includegraphics[width=.97\linewidth]{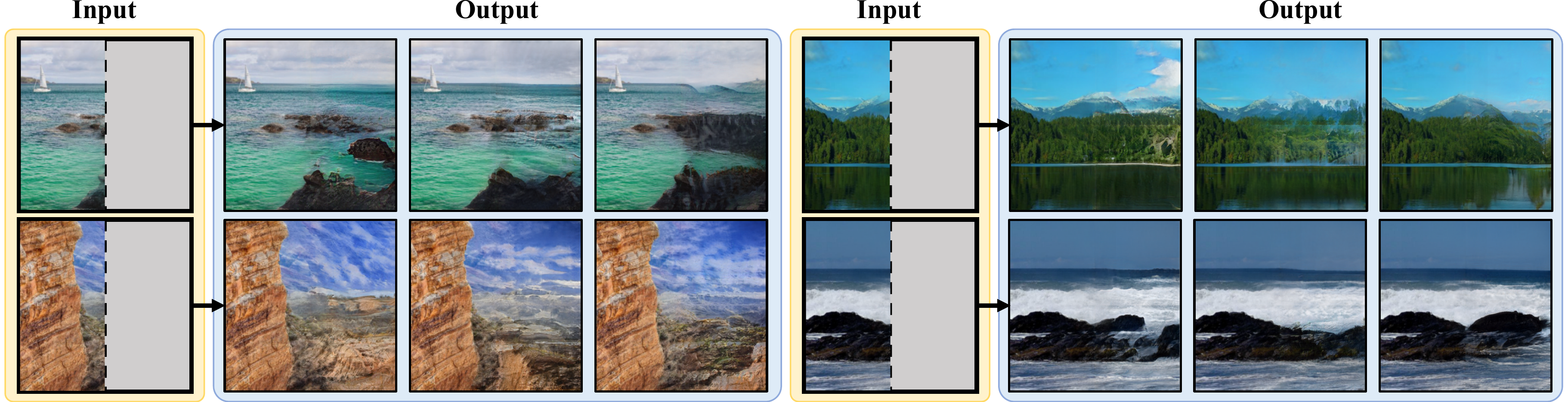}
    \vspace{-2mm}
    \caption{
    \textbf{Diverse outpainting.}
    We show that the proposed method can seek various solutions for a given input, achieving a high-variety of outpainting results without sacrificing the generation quality.
    }
    \vspace \figmargin
    \label{fig:diversity}
\end{figure*}
\begin{figure*}[t]
    \centering
    \includegraphics[width=.97\linewidth]{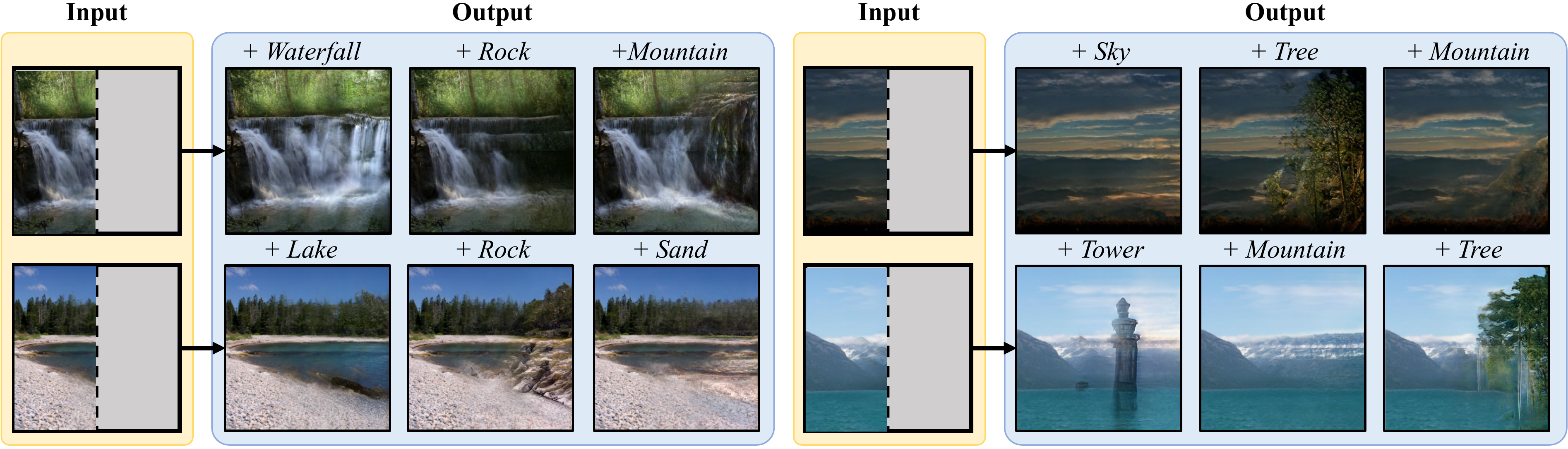}
    \vspace{-2mm}
    \caption{
    \textbf{Categorical generation.}
    We show the effectiveness of categorical manipulation by assigning different categorical labels to the outpainting area of the same real-image input. 
    The results demonstrate that the proposed method can smoothly impose novel objects and calibrate the landscape to accommodate different categorical controls from users.
    }
    \vspace \figmargin
    \vspace{-1mm}
    \label{fig:category}
\end{figure*}

\subsection{Quantitative Evaluation}
\label{subsec:quan}
\vspace{\subsecmargin}

%
We evaluate the results in terms of realism and diversity using the Fr\'echet Inception Distance (FID)~\cite{fid} and Inception Score (IS)~\cite{salimans2016improved}. 
%
%
%
Note that we \textit{do not} apply the blending scheme introduced in~\subsecref{blending} across the quantitative evaluations, as we aim to demonstrate the strength of the proposed pipeline without additional postprocessing.

As shown in \tabref{quan_fid_div}, all of the proposed inversion-based methods perform favorably against I2I-based methods.
The FID and IS results demonstrate that the generated-image distributions from our \ours variants to the real distribution are significantly more similar than the generation distributions from I2I-based baseline methods.
Furthermore, compared to Image2stylegan++~\cite{abdal2020image2stylegan++}, 
%
%
the results show that coordinate conditioning not only enables the categorical manipulation feature, but also naturally improves the generation diversity and quality of Image2stylegan++. 
%
%

\noindent \textbf{User studies.} We conduct user studies to make explicit pairwise qualitative comparisons in two settings: (a) our method against each of the baseline methods, and (b) all methods against real samples.
For each round of comparison, we present a pair of two outpainting results generated from the same real sample to the users. The images are either sampled from our method, baselines, or real images. 
Then, the subjects are asked to select a more realistic and preferred sample out of the image pair. 
We collect the results from $80$ volunteers. Each of them makes $21$ rounds of selections, resulting in $1{,}680$ data points. 

\figref{userstudy} shows that subjects prefer the outpainting results by our model than those by other evaluated methods. 
Especially in comparison with real images, we observe a noticeable gap between the proposed \ours and Boundless. 
This may be attributed to that our method can frequently synthesize complex structures and novel objects (as shown in \figref{comparison} and \figref{category}) that match the varieties and details of real images, while Boundless tends to create overly-smoothed results with raindrop-shaped artifacts.


\vspace{\paramargin}

\begin{figure*}[h!]
    \centering
    \includegraphics[width=.97\linewidth]{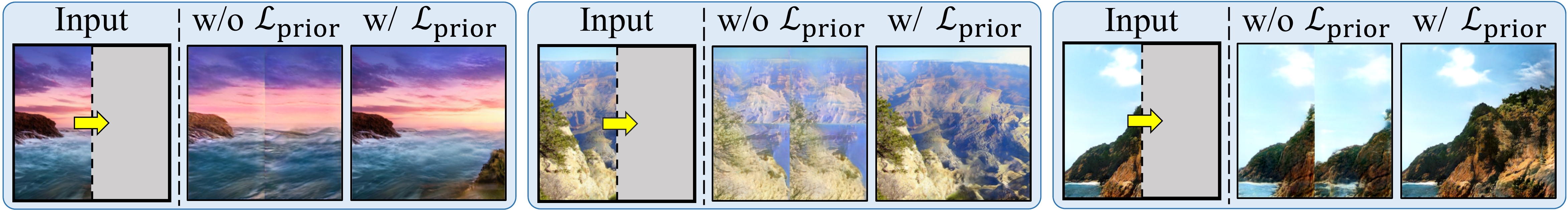}
    \vspace{-2mm}
    \caption{
    \textbf{Qualitative on $\mathcal{L}_{\mathrm{prior}}$.}
    Without $\mathcal{L}_{\mathrm{prior}}$, the inversion will overfit to the reconstruction loss, resulting in latent codes extremely far away from the training distribution. 
    The outpainted area may result in obvious seams (left) or replication of the input image (middle and right).
    %
    }

    \vspace \figmargin
    \vspace{-4mm}
    \label{fig:gaussian_qual}
\end{figure*}

\noindent\textbf{Ablation studies.}
In \tabref{ablation}, we analyze and quantify the effectiveness of each component with ablation studies. 
We introduce a diversity score that measures the perceptual distance~\cite{zhang2018lpips} among $m$ outpainting results with respect to a real image. 
Each diversity score is averaged over $2{,}048$ samples in the testing set. 
%
The diversity loss functions $\mathcal{L}_\mathrm{ms}$ and $\mathcal{L}_\mathrm{div}$ improve the diversity without compromising the quality.
Note that the worst-case diversity quantity without any diversity loss is unlikely to be zero. 
This is due to the stochastic nature of gradient descent and randomization techniques introduced in~\cite{Karras2019stylegan2} for encouraging the exploration during optimization.
%
Furthermore, we show that the prior loss $\mathcal{L}_\mathrm{prior}$ is essential for securing the visual quality.
As we have discussed in \subsecref{inout}, the $\mathcal{L}_\mathrm{prior}$  regularizes the final state of the inverted latent codes to be located within the dense area of the Gaussian prior.
Hence, the generated contents within the outpainted area remain realistic, instead of creating artifacts with the unconstrained latent codes that drift far away from the training distribution.
As shown in \figref{gaussian_qual}, inversion without $\mathcal{L}_\mathrm{prior}$ resulting in either obvious seams between input regions and outpainted areas or replicating input regions to the outpainted areas.
%
%
%
%
%

We evaluate the effect of different $m$ values of $\mathcal{L}_\mathrm{ms}$ and $\mathcal{L}_\mathrm{div}$.
%
We demonstrate that the diversity losses provide significant improvement in diversity score without compromising visual quality by seeking distinctive latent codes.

\begin{figure*}[t!]
    \centering
    \includegraphics[width=\linewidth]{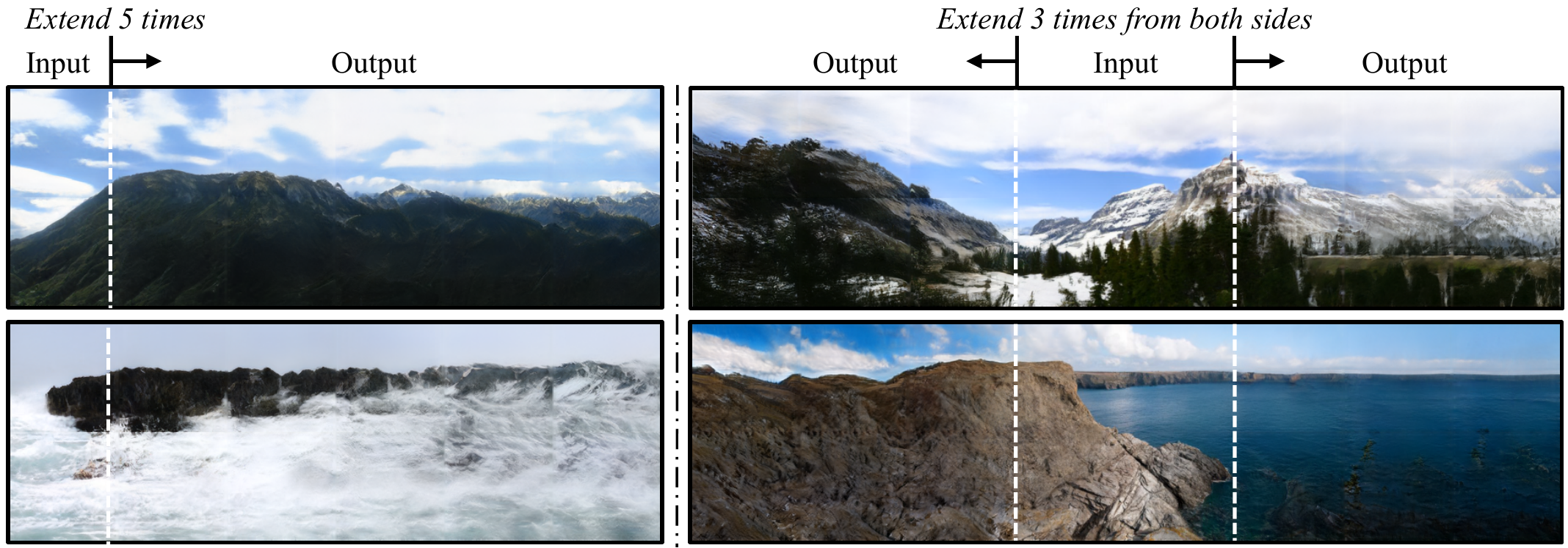}
    \vspace{-7mm}
    \caption{
    \textbf{Panorama generation.} We synthesize panoramic images by performing recursive outpainting. The results are of high quality and high structural complexity without repeating patterns.
    }
    \vspace \figmargin
    \vspace{-0mm}
    \label{fig:panorama}
\end{figure*}

\begin{figure}[h!]
    \centering
    \includegraphics[width=\linewidth]{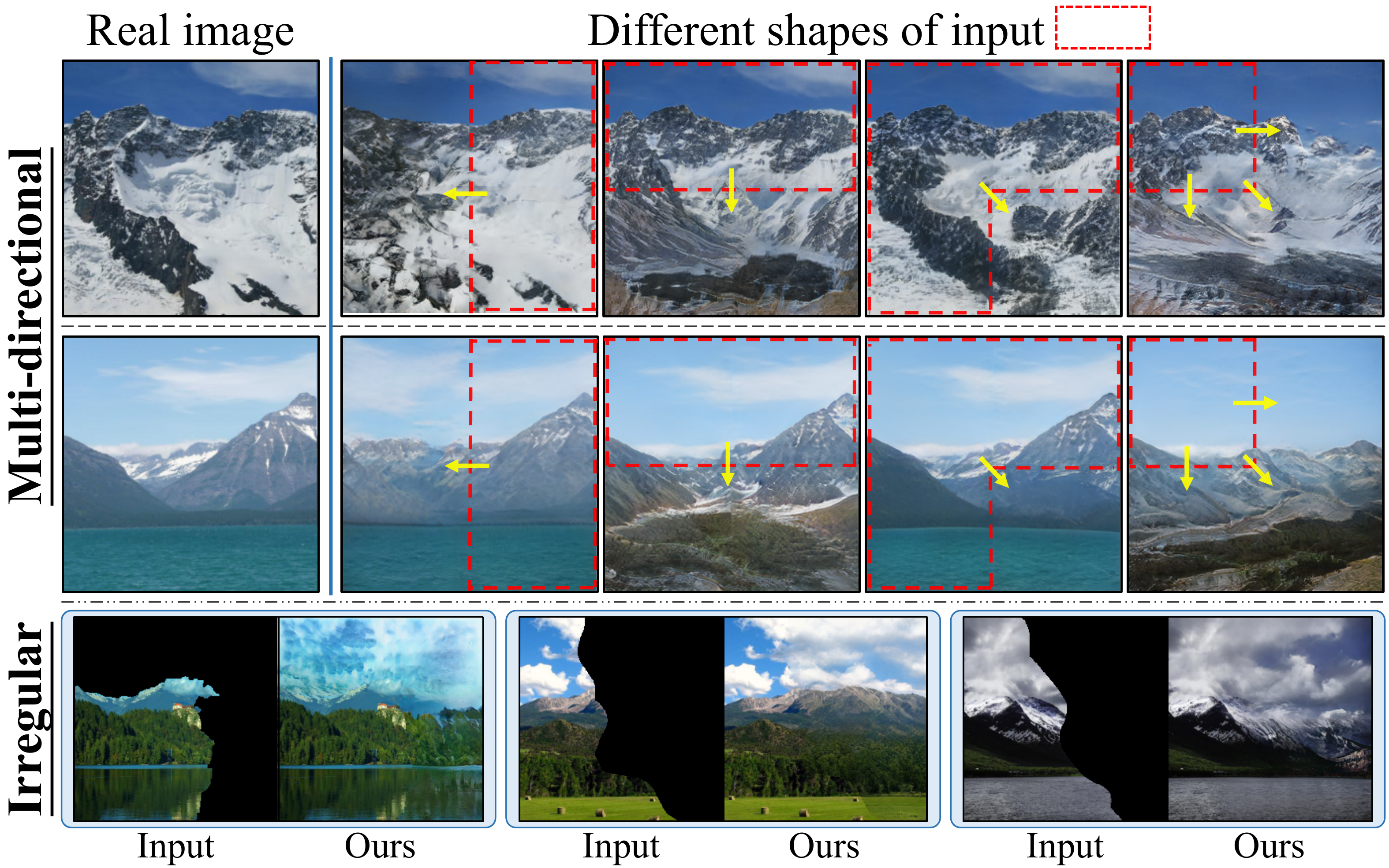}
    \vspace{-6mm}
    \caption{
    \textbf{Multi-directional and irregular boundary outpainting.}
    Our pipeline can inherently tackle \textit{(top)} different outpainting directions and \textit{(bottom)} irregular input shapes.
    }
    \vspace \figmargin
    \vspace{-2mm}
    \label{fig:multi-directional}
\end{figure}

\vspace{\subsecmargin}
\subsection{Qualitative Evaluation}
\label{subsec:qual}
\vspace{\subsecmargin}
In this section, we demonstrate the visual quality and diversity of the proposed method, and present applications, including categorical generation, panorama generation, and outpainting from different shapes and directions.
Please refer to the supplementary materials for more visual results.

\vspace{\paramargin}

\noindent\textbf{Visual quality.} 
In \figref{comparison}, we compare the visual quality of outpainting results from all methods
%
The results show that \ours is generally more realistic, coherent, diverse, exhibits more novel structures/objects, yet introduces fewer noticeable artifacts. 
In contrast, Boundless~\cite{teterwak2019boundless} tends to introduce raindrop-shaped artifacts, DeepFillv2~\cite{yu2018generative,yu2018free} creates blurry extensions, while NS-outpaint~\cite{yang2019very} and Image2stylegan++~\cite{abdal2020image2stylegan++} frequently generates strong artifacts and obvious color differences.
%
\vspace{\paramargin}

\noindent\textbf{Diverse outpainting.} In \figref{diversity}, we show the diverse outpainting results when $m=3$. 
The results show that the proposed diversity loss enables the inversion pipeline to seek for different outpainting solutions.
Notice that despite the variety of outpainting solutions, all inverted results remain visually compelling and match the real-image input.

\noindent\textbf{Categorical Generation.}
\figref{category} shows the results of categorical manipulation enabled with the \oursc~variant.
Users can insert class-specific objects or manipulate the outpainted landscape structure with the categorical conditions of the two micro-patches on the right. 
The generator is able to automatically complete the background as well as blending the presented objects into the scene.

\noindent\textbf{Panorama generation.}
Our framework naturally supports panorama generation by recursively taking previous outpainted micro-patches as the new inversion target. 
\figref{panorama} shows the panoramas generated from our method by outpainting to the left and right. 
The results show that the recursively outpainted area contains highly diverse structures without repeating patterns.

\noindent\textbf{Multi-directional and irregular-boundary outpainting.}
For brevity, most results presented are generated given two micro-patches on the left-hand side as inputs. 
Nevertheless, the proposed method can perform outpainting from various directions with different input shapes and even arbitrary input shapes.
In the top of \figref{multi-directional}, from left to right, we demonstrate outpainting results generated from the right, from the top, given three micro-patches, and given one micro-patch.
In the bottom of \figref{multi-directional}, we present outpainted results given inputs of irregular boundaries.


\vspace{-0mm}
\section{Conclusion}
\vspace{\secmargin}
\label{sec:conclusion}
In this work, we tackle the image outpainting task from the GAN inversion perspective.
%
We first train a generator to synthesize micro-patches conditioned on their positions.
Based on the trained generator, we propose an inversion process that seeks for multiple latent codes recovering available regions as well as predicting outpainting regions.
The proposed framework can generate diverse samples and support categorical specific outpainting, enabling more flexible user controls.
Qualitative and quantitative experiments demonstrate the effectiveness of the proposed framework in terms of visual quality and diversity. 

\clearpage
{\small
\bibliographystyle{ieee_fullname}
\bibliography{a-citation}
}

\clearpage

\begin{appendix}

\section{Overview}
\vspace{-2mm}
In this supplementary materials, we describe the detail of the datasets we used in \secref{data}.
We investigate the failure cases and limitation of the proposed model in \secref{fail} and provide more qualitative results in \secref{qual_results}.
We also analyze the effect of applying the Gaussianized Space to the proposed method in \secref{gaussian}.

\section{Dataset}
\vspace{-2mm}
\label{sec:data}
For the Places365 dataset, we pick the following categories to form the training set: \textit{bamboo, forest, beach, bridge, canyon, cliff, corn field, dam, desert, farm, field, forest, path, glacier, hayfield, hot spring, lake, mountain, ocean, rainforest, snowfield, valley, volcano, waterfall, wave}.

For the Flickr-Scenery dataset, we construct the dataset by manually searching for and crawling images with the following keywords: \textit{aurora, beach, bridge, canyon, cliff, forest, fountain, glacier, hayfield, lake, lighthouse, maple, meteor shower, mountain, ocean, sakura, snowfield, storm, sunrise, sunset, valley, waterfall, wave, wisteria}.

\vspace{1mm}
\noindent \textbf{Categorical setting.} To facilitate the model training without the burden of massive categories, for both datasets, we train our model with a subset of categories related to scenery and with sufficient amount of samples.
These categories are: \textit{sky, tree, road, grass, sidewalk, earth, mountain, plant, water, sea, field, rock, sand, skyscraper, path, runway, river, bridge, hill, tree, light, tower, dirt, land, stage, fountain, pool, waterfall, lake, pier}.
To avoid the situation that any particular class occupies only a negligibly small region in the image that poses difficulties to the classifier training, we consider the pixels as background if a class presented in the micro-patch covers less than $1\%$ of the area.
%
%


\section{Additional Results}
\begin{figure*}[h!]
    \centering
    \includegraphics[width=\linewidth]{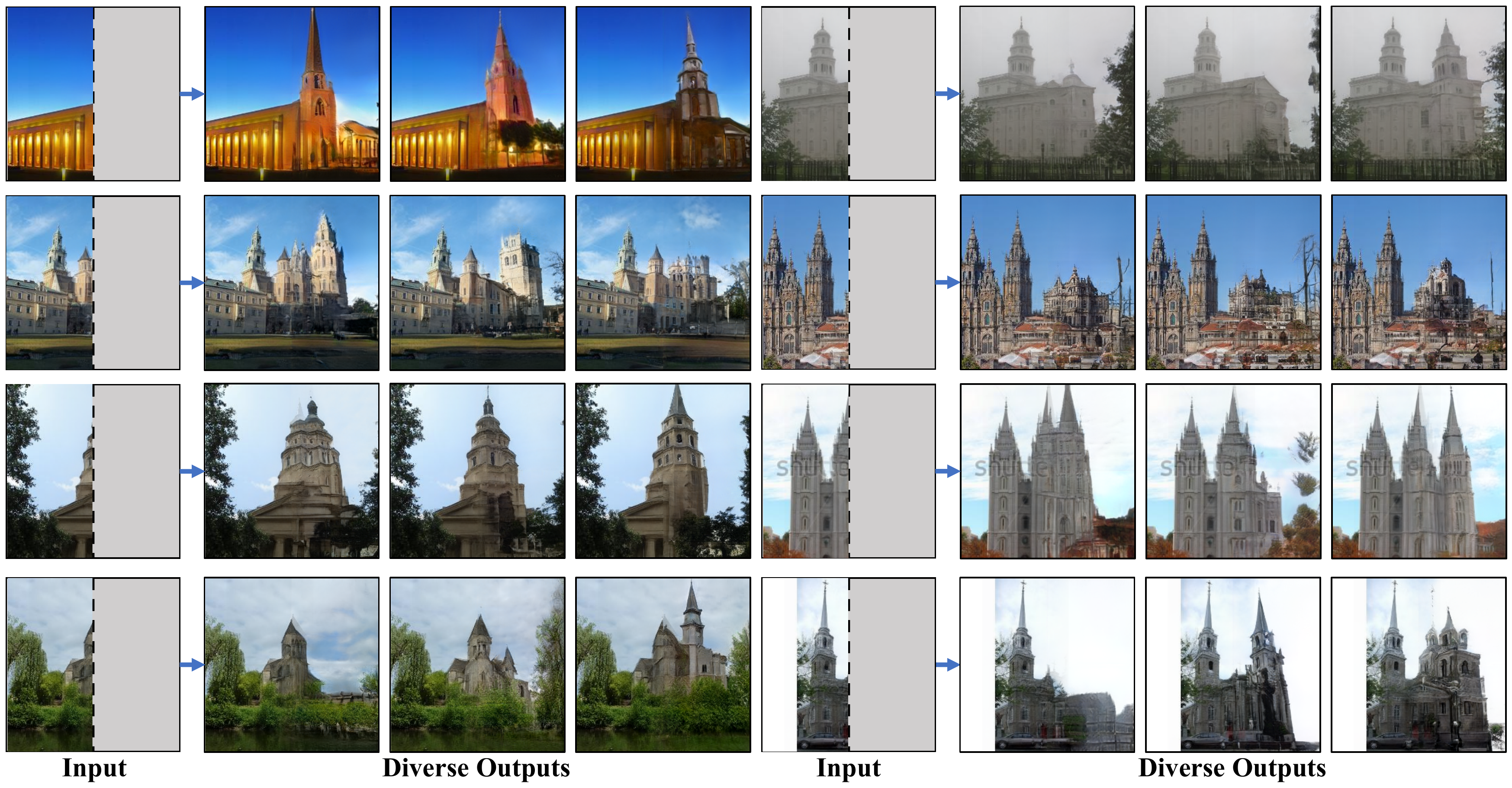}
    \caption{
    \textbf{Qualitative results on LSUN church.} We show that the proposed method could handle datasets with structured objects as well.
    }
    \vspace \figmargin
    \label{fig:church}
\end{figure*}

\begin{figure*}[t!]
    \centering
    \includegraphics[width=\linewidth]{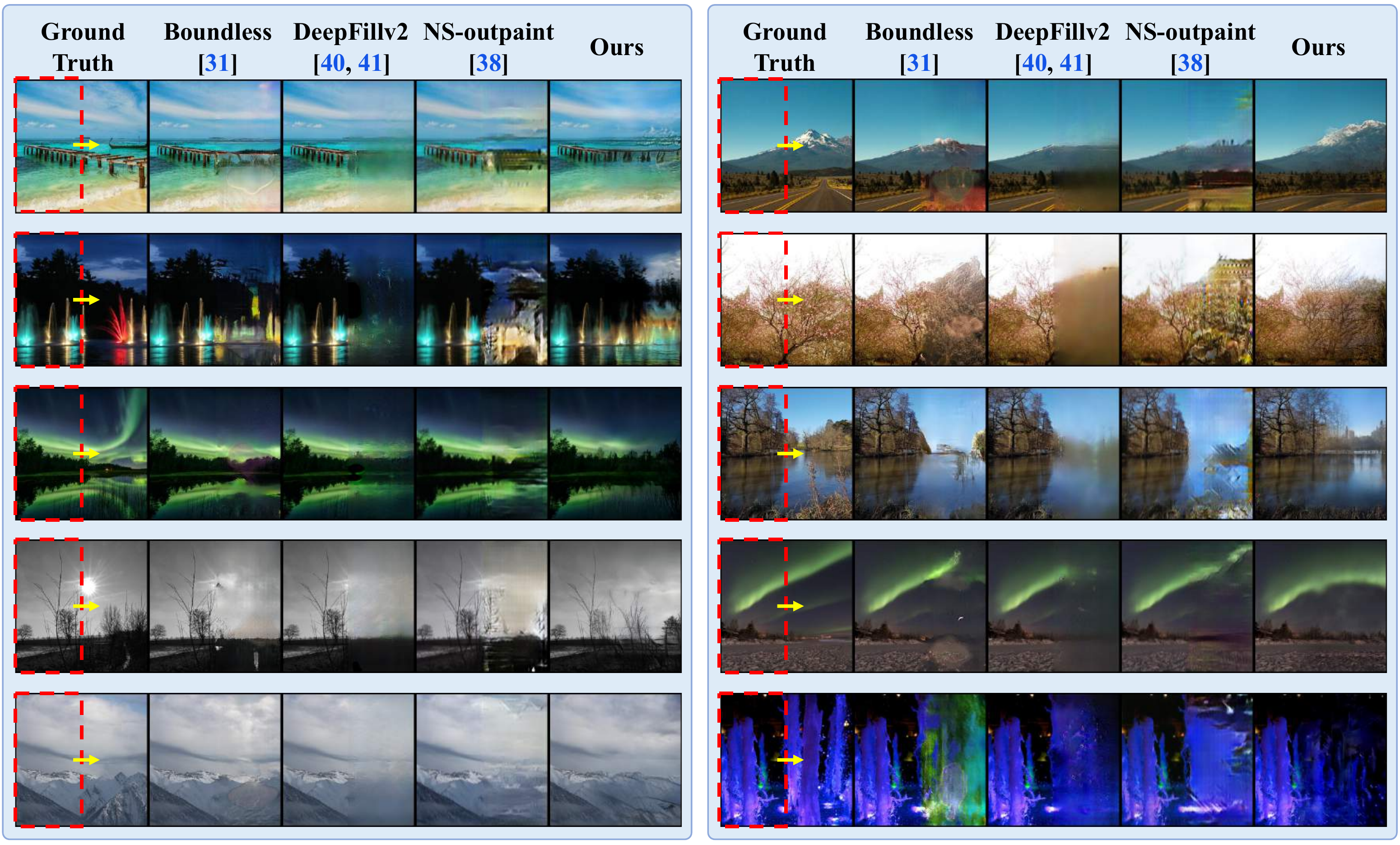}
    \caption{
    \textbf{More comparison to related work.} 
    The qualitative comparison against other related methods show that the proposed approach is more stable, synthesizes richer context with more complex structures, and is able to handle some of the difficult complex scenes.
    (The input regions to all methods are marked with red dashes.)
    }
    \vspace \figmargin
    \label{fig:more_comp}
\end{figure*}

\begin{figure*}[t!]
    \centering
    \includegraphics[width=\linewidth]{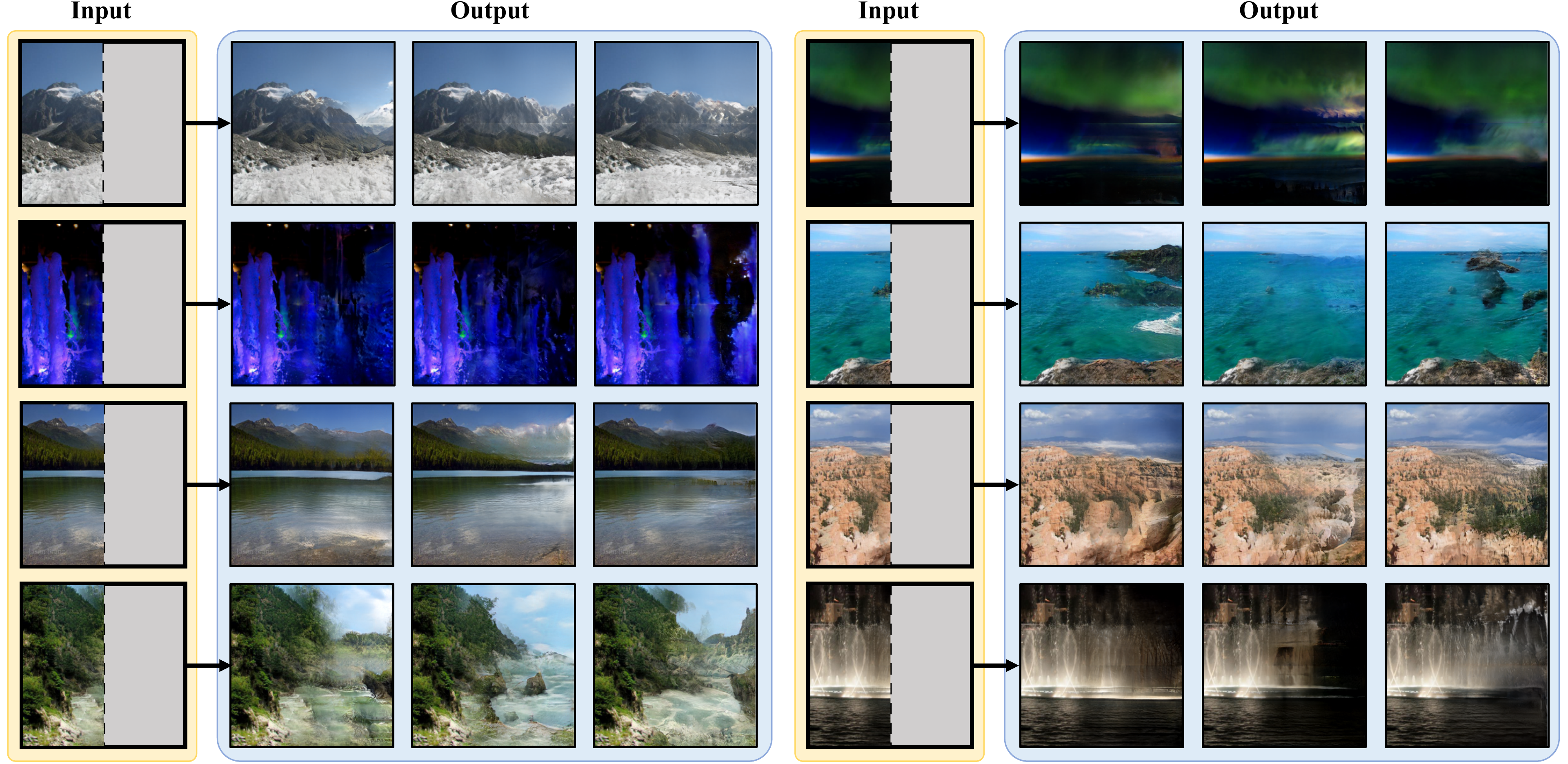}
    \caption{
    %
    \textbf{More results on diverse outpainting.} 
    We show that the proposed method can seek various solutions for a given input, achieving a high-variety of outpainting results without sacrificing the generation quality.
    }
    \vspace \figmargin
    \label{fig:more_diverse}
\end{figure*}

\begin{figure*}[t!]
    \centering
    \includegraphics[width=\linewidth]{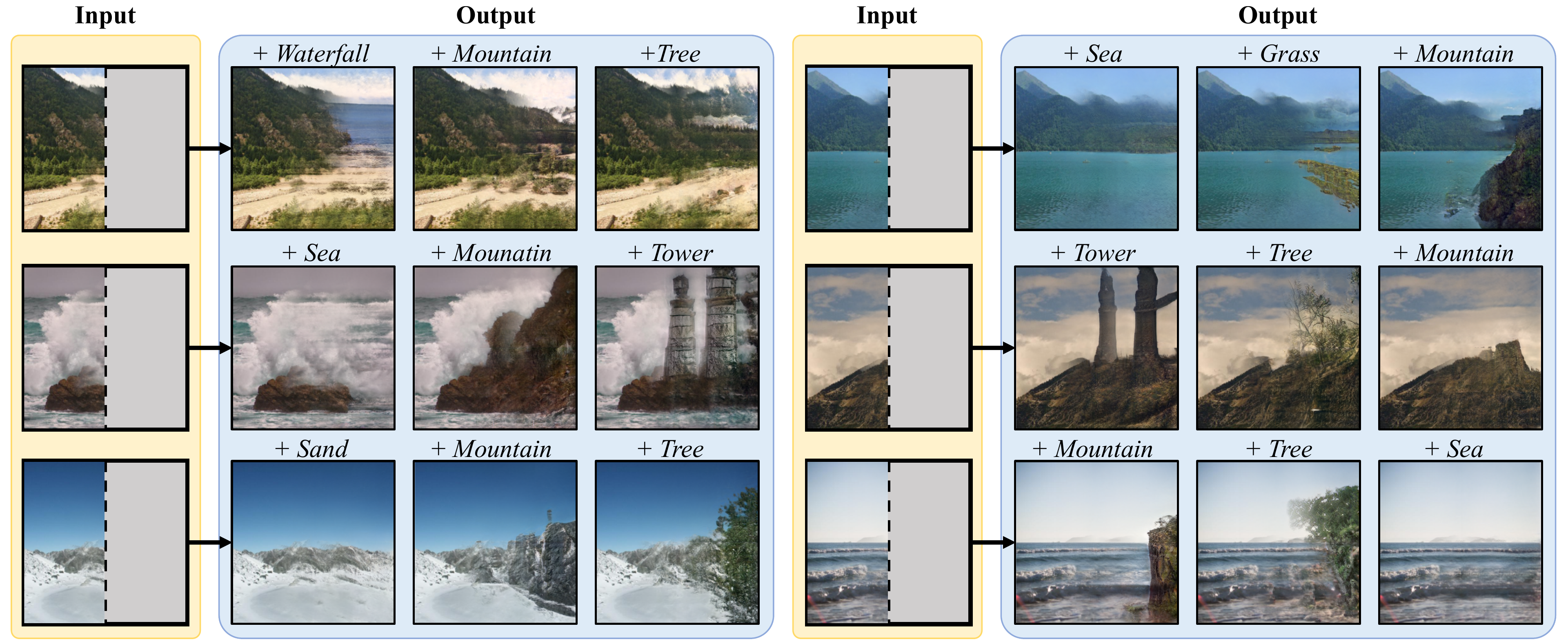}
    \caption{
    %
    \textbf{More categorical inversion results.} We show the effectiveness of categorical manipulation by assigning different categorical labels to the outpainting area of the same real-image input. 
    The results demonstrate that the proposed method can smoothly impose novel objects and calibrate the landscape to accommodate different categorical controls from users.
    }
    \vspace \figmargin
    \label{fig:more_cat}
\end{figure*}

\begin{figure*}[t!]
    \centering
    \includegraphics[width=\linewidth]{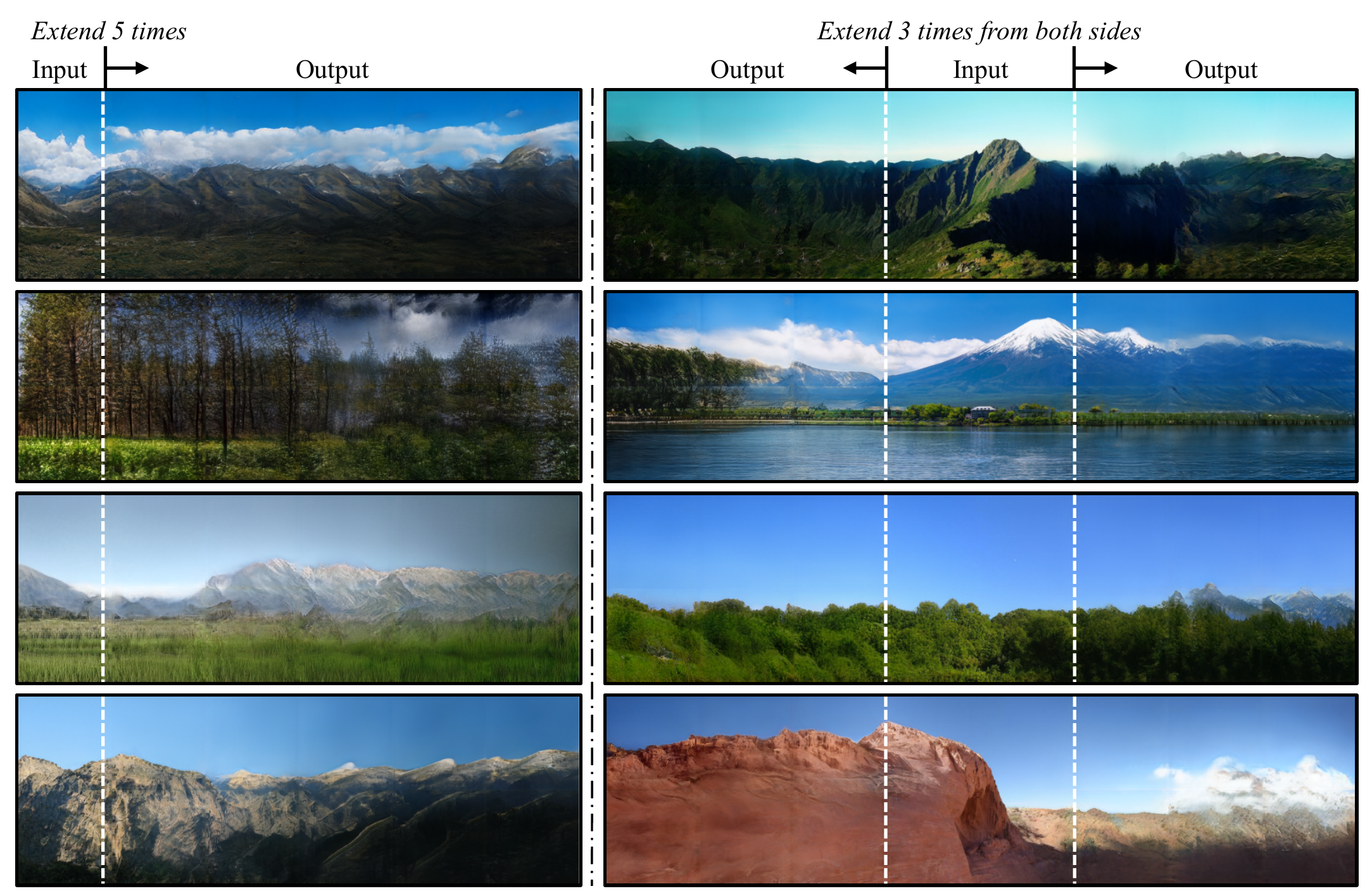}
    \caption{
    %
    \textbf{More results on panorama generation.} 
    We synthesize panoramic images by performing recursive outpainting. The results are of high quality and high structural complexity without repeating patterns.
    }
    \vspace \figmargin
    \label{fig:more_pano}
\end{figure*}

\vspace{-2mm}
\label{sec:qual_results}
First, to demonstrate the generalizability of the proposed method, we present the results on the LSUN Church~\cite{yu2015lsun} dataset. As shown in \figref{church}, the proposed method can be applied to images with structural and artificial contents in addition to landscape images.
Then we present 1) more qualitative comparisons with the baselines in \figref{more_comp}, 2) more multimodal generation results in \figref{more_diverse}, 3) more categorical manipulation results in \figref{more_cat}, and 5) more panorama generation results in \figref{more_pano}.

\section{Failure Cases and Limitation}
\begin{figure*}[h!]
    \centering
    \includegraphics[width=.8\linewidth]{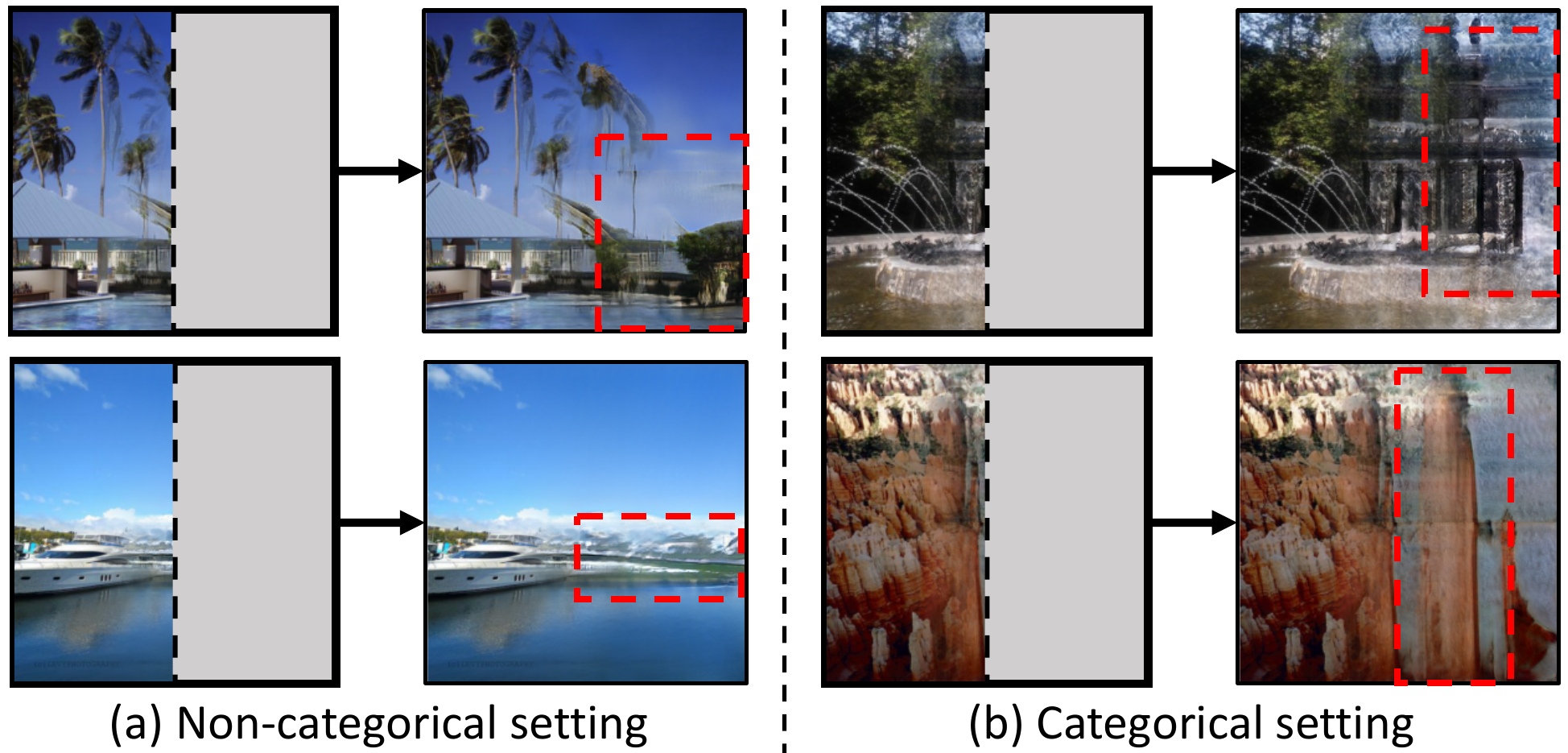}
    \caption{
    \textbf{Failure Cases.} We show some limitations of the proposed method.
    First, the input is out of distribution. For example, the coconut tree and the villa at the first row, and the yacht at the second row of \figref{fail}(a). Similarly, the fountain basin at the first row of \figref{fail}(b).
    Second, the unseen category combination in the categorical setting. For instance, trying to generate tower in the unknown area while the known region is the valley in the second row of \figref{fail}(b).
    }
    \vspace \figmargin
    \label{fig:fail}
\end{figure*}

\vspace{-2mm}
\label{sec:fail}
Despite the success in producing more complex and visually plausible structures in the outpainting area, we still observe some limitations that are intriguing for future study. The failure part is indicated by the red dashed boxes in \figref{fail}.
%
First, the input is out of distribution. In this case, the latent space is not fully explored and we cannot find an proper latent code to outpaint the missing region. For example, in the non-categorical setting, the coconut tree and the villa at the first row, and the yacht at the second row of \figref{fail}(a). Similarly, in the categorical setting, the fountain basin at the first row of \figref{fail}(b).
Second, the unseen category combination in the categorical setting. For categorical manipulation, if the outpainted region is assigned with a category which is rarely appeared together with the original category, the outpainted results will fail in this case. For instance, trying to generate tower in the unknown area while the known region is the valley in the second row of \figref{fail}(b).
%

%
%
%
%
%

\section{Ablation on Gaussianized Space}
\begin{figure*}
    \centering
    \includegraphics[width=\linewidth]{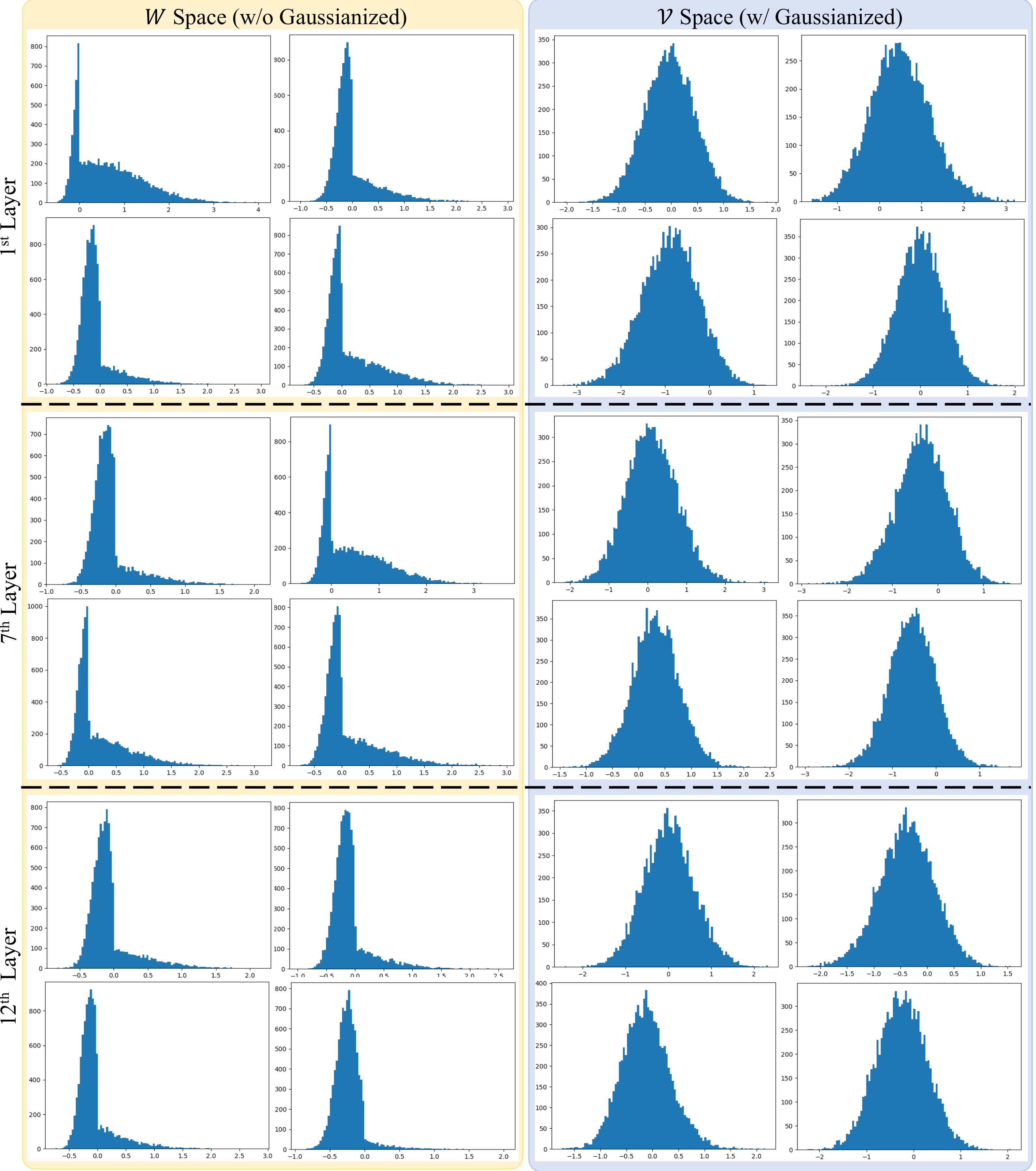}
    \caption{
    \textbf{Ablation on Gaussianized space.} For the feature values in the $1^{st}$, $7^{th}$, $12^{th}$ layer of the generator, we randomly sample four different channels, collapse the feature into an one-dimensional vector, and visualize the values with histograms. We show that the distributions are significantly reshaped into a Gaussian-like distribution after applying the Gaussianized space $\mathcal{V}$.
    }
    \vspace \figmargin
    \label{fig:prior_ablation}
\end{figure*}

\vspace{-2mm}
\label{sec:gaussian}
As described in the Section 3.2 of the main paper, it is crucial to encourage the target latent code to belong to the training distribution, rather than overfitting to the given image and resulting in an extremely out-of-domain latent code. 
Here we empirically show the differences in distribution between the $\mathcal{W}$ space and Gaussianized space $\mathcal{V}$ in \figref{prior_ablation}.
We plot the histogram of latent codes from the $1^{st}$, $7^{th}$, $12^{th}$ layer of the trained generator.
We can observe that the latent codes sampled from the $\mathcal{V}$ space are more aligned to the Gaussian distribution.
This behavior largely constrains the search space during the inversion process, and significantly improves the visual quality.


\clearpage


\end{appendix}

\end{document}